\documentclass[conference,11pt]{IEEEtran}
\IEEEoverridecommandlockouts
\usepackage[letterpaper,top=0.75in,bottom=1.08in,left=0.68in,right=0.68in,columnsep=0.28in]{geometry}
\usepackage{url}
\usepackage[font=small,skip=3pt]{caption}
\usepackage{cite}
\usepackage{amsmath,amssymb,amsfonts}
\usepackage{algorithm}
\usepackage{algpseudocode}
\usepackage{graphicx}
\usepackage{booktabs}
\usepackage{textcomp}
\usepackage{xcolor}
\usepackage[colorlinks=true,allcolors=blue]{hyperref}
\def\BibTeX{{\rm B\kern-.05em{\sc i\kern-.025em b}\kern-.08em
    T\kern-.1667em\lower.7ex\hbox{E}\kern-.125emX}}

\algnewcommand\algorithmicforeach{\textbf{for each}}
\algdef{S}[FOR]{ForEach}[1]{\algorithmicforeach\ #1\ \algorithmicdo}

\begin{document}

\title{APGEM: Adaptive Policy-Guided Error Mitigation for Quantum Reinforcement Learning on a Real-World CVRP Case Study\\}

\author{%
\IEEEauthorblockN{}%
\setlength{\tabcolsep}{2pt}%
{\small
\begin{tabular*}{\textwidth}{@{\extracolsep{\fill}} c c c @{}}
\begin{tabular}[t]{@{}c@{}}
Shabir Ahmad Sofi\textsuperscript{*}\\
\textit{Department of Information Technology}\\
\textit{NIT Srinagar}\\
J\&K, India\\
shabir@nitsri.ac.in
\end{tabular}
&
\begin{tabular}[t]{@{}c@{}}
Bisma Majid\textsuperscript{*}\\
\textit{Department of Information Technology}\\
\textit{NIT Srinagar}\\
J\&K, India\\
bismabhat\_ite006@nitsri.ac.in
\end{tabular}
&
\begin{tabular}[t]{@{}c@{}}
Mir Mohammad Yousuf\textsuperscript{*}\\
\textit{Department of Information Technology}\\
\textit{NIT Srinagar}\\
J\&K, India\\
yousuf\_2022phaite006@nitsri.ac.in
\end{tabular}
\end{tabular*}}%
\thanks{*These authors contributed equally to this work.}%
}

\maketitle

\begin{abstract}
Quantum Reinforcement Learning (QRL) represents policies as variational quantum circuits (VQCs), making it attractive for combinatorial optimization such as the Capacitated Vehicle Routing Problem (CVRP). On noisy intermediate-scale quantum (NISQ) hardware, however, decoherence degrades fidelity and destabilizes learning, and conventional error mitigation is applied statically without regard to the learning context. We introduce Adaptive Policy-Guided Error Mitigation (APGEM), a controller that selects among Zero-Noise Extrapolation (ZNE), Probabilistic Error Cancellation (PEC), Clifford Data Regression (CDR), and Readout Error Mitigation (REM) online, driven by a fidelity, entropy, and cost aware utility function and an $\varepsilon$-greedy rule over temporal-difference Q-scores. We evaluate on a realistic urban-logistics testbed, a Delhi-based CVRP over real landmarks with geodesic inter-node costs, exercised across five noise families and four severity levels. On this instance, the QRL agent outperforms constructive heuristics and approaches metaheuristics, while mitigation restores approximation ratios from $0.84$--$0.87$ to $0.92$--$0.94$ under high noise. The controller shifts from a CDR-dominated regime under short training horizons to a balanced deployment across all four techniques under longer horizons, indicating genuine regime-dependent selection. These preliminary results position adaptive, learning-aware mitigation as a practical route to noise-resilient QRL.
\end{abstract}


\section{Introduction}
The Vehicle Routing Problem (VRP) is a canonical NP-hard combinatorial optimization task underpinning transportation and last-mile logistics \cite{vidal2022hybrid,mor2022vehicle,ghosal2024unifying}. Efficient routing yields direct savings in cost, fuel, and emissions, so the problem is a persistent driver of efficiency in industry \cite{sabet2022green,los2022large}. Classical exact solvers and metaheuristics scale poorly as the number of customers and constraints grows, motivating alternative computational paradigms \cite{ochelska2021selected,bouanane2022vehicle}. Noisy intermediate-scale quantum (NISQ) devices have emerged as one such candidate \cite{preskill2023quantum,bharti2022noisy}, with variational quantum algorithms particularly suited to near-term hardware \cite{cerezo2021variational}.

Quantum Reinforcement Learning (QRL) embeds parameterized quantum circuits within reinforcement learning agents, representing the policy as a variational quantum circuit (VQC) that maps environment states to action probabilities while classical optimization updates the parameters \cite{kolle2024study}. This hybrid structure suits sequential decision problems such as VRP, and recent work applies parameterized quantum policies directly to routing and last-mile delivery \cite{moosavi2025quantum}. Two obstacles limit QRL in practice. First, circuits executed on NISQ devices are highly susceptible to noise, which lowers fidelity, raises state entropy, and destabilizes learning \cite{kim2023scalable}. Second, established mitigation methods, including Zero-Noise Extrapolation (ZNE), Probabilistic Error Cancellation (PEC), Clifford Data Regression (CDR), and Readout Error Mitigation (REM), are typically chosen once and applied uniformly, without adapting to the circuit or noise characteristics of a given execution \cite{pelofske2025digital,nation2021scalable,ding2022evaluating}. Learning-based mitigation has begun to close this gap for variational circuits and quantum software \cite{liao2024machine,muqeet2024machine, yousuf2026systematic}, yet it is rarely coupled to the reinforcement signal that drives policy optimization. How noise-aware adaptation should influence policy optimization, and how to select among mitigation strategies in real time from observed performance signals, remains open \cite{czarnik2021error}.

We address this gap with \textbf{Adaptive Policy-Guided Error Mitigation (APGEM)}. Our contributions are:
\begin{enumerate}
    \item A hybrid QRL environment instantiated on a real-world CVRP case study, a Delhi last-mile delivery scenario with authentic geographic structure, in which the policy is a VQC trained under noise models that emulate realistic NISQ conditions.
    \item The APGEM controller, which dynamically selects among ZNE, PEC, CDR, and REM using an $\varepsilon$-greedy rule guided by a composite utility over fidelity, entropy, and computational cost, turning mitigation from a static pre-processing step into a learning component coupled to reinforcement feedback.
    \item An adaptive learning-rate scheme that modulates parameter updates by both reward and quantum execution quality, stabilizing training under noise.
\end{enumerate}

\section{Related Work}
\textbf{Quantum computing for logistics.} Quantum and hybrid quantum-classical methods have been explored across supply-chain and routing tasks, motivated by the combinatorial hardness of VRP variants and the economic value of marginal routing improvements \cite{fakhravar2022combining,ambrosino2022rich,sabet2022green}. These efforts largely target the optimization formulation itself, for example through annealing or QAOA-style objectives, rather than through learned sequential policies.

\textbf{Quantum reinforcement learning.} QRL replaces classical policy or value networks with variational quantum circuits, exploiting quantum state representation for compact policy encodings \cite{kolle2024study,majid2025quantum,lamichhane2025quantum}. Recent studies apply parameterized quantum policies to routing and last-mile delivery under realistic constraints \cite{moosavi2025quantum}, but typically assume ideal or lightly modeled noise and do not treat error mitigation as part of the learning loop.

\textbf{Quantum error mitigation.} A mature toolbox of mitigation techniques now exists, including ZNE \cite{giurgica2020digital,mari2021extending}, readout correction \cite{nation2021scalable}, and calibration-based regression \cite{czarnik2021error}, surveyed comprehensively in \cite{cai2023quantum}. More recent work learns mitigation models directly from data for variational circuits and quantum software pipelines \cite{liao2024machine,muqeet2024machine}. These methods are powerful but are almost always applied as a fixed, pre-selected stage. Our work differs by making the choice of mitigation technique itself an online, reward-coupled decision within a QRL agent.

\section{Background}
\subsection{CVRP as a Reinforcement Learning Problem}
The CVRP seeks minimum-cost tours for a fleet of capacity-limited vehicles serving customers from a single depot. We cast it as an episodic Markov decision process. The state encodes vehicle positions, remaining loads, partial route lengths, and per-customer visit and remaining-demand indicators. The action set $\mathcal{A}(s_t)$ comprises unserved customers within remaining capacity plus a return-to-depot action. The step reward combines a distance term with a visit bonus, and a terminal reward or penalty reflects unserved customers.

\subsection{Quantum Reinforcement Learning Formulation}
The policy $\pi_\theta$ is realized by a variational quantum circuit $U(\theta)$ with trainable parameters $\theta=(\theta_1,\dots,\theta_p)$. Acting on $n_q$ qubits it prepares
\begin{equation}
|\psi(\theta)\rangle = U(\theta)\,|0\rangle^{\otimes n_q},
\end{equation}
and a computational-basis measurement yields a bitstring $z$ with probability $P_\theta(z)=|\langle z|\psi(\theta)\rangle|^2$. A classical post-processing map $a_t=f(z,s_t)$ enforces feasibility, so the induced policy is
\begin{equation}
\pi_\theta(a_t \mid s_t) = \sum_{z \in \mathcal{Z}(a_t)} P_\theta(z),
\end{equation}
where $\mathcal{Z}(a_t)$ is the set of bitstrings mapped to action $a_t$. Training maximizes the expected return $J(\theta)=\mathbb{E}_{\pi_\theta}[\sum_{t=0}^T r_t]$. Since the reward encodes negative travel cost, maximizing $J(\theta)$ is equivalent to minimizing the classical routing cost $\sum_{k \in K}\sum_{(i,j)\in \text{route}_k} c_{ij}$. Gradients use the parameter-shift rule,
\begin{equation}
\frac{\partial}{\partial \theta_i}\langle O\rangle_\theta = \tfrac{1}{2}\!\left(\langle O\rangle_{\theta_i+\frac{\pi}{2}} - \langle O\rangle_{\theta_i-\frac{\pi}{2}}\right),
\end{equation}
enabling unbiased stochastic gradient ascent $\theta \leftarrow \theta + \eta\,\nabla_\theta J(\theta)$.

\subsection{Noise Modeling}
On near-term devices the circuit executes under a completely positive trace-preserving map $\mathcal{N}$, so the prepared state is $\rho(\theta)=\mathcal{N}(U(\theta)|0\rangle\langle 0|^{\otimes n_q}U^\dagger(\theta))$ with Kraus form $\mathcal{N}(\rho)=\sum_\alpha K_\alpha \rho K_\alpha^\dagger$ and $\sum_\alpha K_\alpha^\dagger K_\alpha = I$. We simulate four canonical channels: the depolarizing channel $\mathcal{D}_p(\rho)=(1-p)\rho+\tfrac{p}{3}(X\rho X+Y\rho Y+Z\rho Z)$; dephasing $\mathcal{Z}_\lambda(\rho)=(1-\lambda)\rho+\lambda Z\rho Z$; amplitude damping with Kraus operators parameterized by $\gamma$; and readout noise modeled by a confusion matrix $A$ with $\tilde{p}=Ap$. Two-qubit gate noise is applied as depolarizing noise on entangling gates.

\subsection{Error Mitigation Techniques}
The mitigation pool spans complementary tradeoffs \cite{cai2023quantum}. ZNE scales the effective noise to levels $\lambda_i$ and extrapolates expectation values to the zero-noise limit, for example by Richardson extrapolation $\hat{E}(0)=\sum_{i=1}^{r} c_i E(\lambda_i)$, reducing bias at the cost of added variance \cite{giurgica2020digital,mari2021extending}. PEC expresses a target gate as a quasi-probability decomposition $\mathcal{G}=\sum_j \alpha_j \tilde{\mathcal{G}}_j$ and reweights outcomes by $\mathrm{sgn}(\alpha_j)\Gamma$ with $\Gamma=\sum_j|\alpha_j|$, giving an unbiased estimate with variance overhead scaling as $\Gamma^2$. CDR fits a regression $y\approx ax+b$ between noisy and ideal expectation values from near-Clifford proxy circuits and benefits from richer calibration data \cite{czarnik2021error}. REM inverts the confusion matrix, $\hat{p}=A^{-1}\tilde{p}$, to correct measurement statistics \cite{nation2021scalable}. No single method dominates across noise regimes or across the training trajectory, which motivates adaptive selection.

\section{Adaptive Policy-Guided Error Mitigation}
The effectiveness of each mitigation method varies over training: early policies are highly stochastic and tolerate low-cost mitigation, whereas sharpened later policies demand more aggressive correction to preserve fidelity. APGEM frames the choice of mitigation as a learning problem guided by policy diagnostics.

For a policy parameter $\theta$ and state $s$, we track the fidelity $F(\theta,s)$ of the mitigated state relative to the ideal, the entropy of the policy distribution,
\begin{equation}
H(\theta,s) = -\sum_z \tilde{p}_\theta(z \mid s)\log \tilde{p}_\theta(z \mid s),
\end{equation}
the variance of corrected estimates, and the computational cost $C_\mathcal{M}$. These are aggregated into a scalar utility for each candidate method $\mathcal{M}$:
\begin{equation}
U(\mathcal{M};\theta,s) = w_F F_\mathcal{M}(\theta,s) - w_H H_\mathcal{M}(\theta,s) - w_C C_\mathcal{M},
\end{equation}
where $w_F, w_H, w_C$ weight fidelity, entropy, and overhead. The controller maintains Q-scores $Q_t(\mathcal{M})$ updated by a temporal-difference rule,
\begin{equation}
Q_{t+1}(\mathcal{M}) = (1-\alpha)Q_t(\mathcal{M}) + \alpha\, U(\mathcal{M};\theta_t,s_t),
\end{equation}
with learning rate $\alpha \in (0,1)$, and selects a method $\varepsilon$-greedily:
\begin{equation}
\mathcal{M}_t =
\begin{cases}
\text{random } \mathcal{M}, & \text{with probability } \varepsilon,\\
\arg\max_{\mathcal{M}} Q_t(\mathcal{M}), & \text{with probability } 1-\varepsilon.
\end{cases}
\end{equation}
The selected method yields corrected statistics $\tilde{p}_\theta^{\mathcal{M}_t}(z \mid s)$ and a mitigated policy $\pi_\theta^{\mathcal{M}_t}$ used for action sampling and gradient estimation, so that the policy gradient
\begin{equation}
\nabla_\theta J(\theta;\mathcal{M}_t) = \mathbb{E}\!\left[\sum_{t=0}^{T} \nabla_\theta \log \pi_\theta^{\mathcal{M}_t}(a_t \mid s_t)\, G_t\right]
\end{equation}
is informed by adaptively mitigated measurements, with $G_t$ the discounted return-to-go. Algorithm~\ref{alg:apgem} summarizes the controller.

\begin{algorithm}[h]
\caption{APGEM Controller}
\label{alg:apgem}
\begin{algorithmic}[1]
\State \textbf{Input:} pool $\mathcal{M}=\{\text{ZNE, PEC, CDR, REM}\}$, rates $\alpha,\varepsilon$
\State \textbf{Init:} $Q(\mathcal{M}) \gets 0\ \forall \mathcal{M}$
\Procedure{APGEM}{$\theta, s_t$}
    \ForEach{$\mathcal{M} \in \mathcal{M}$}
        \State apply $\mathcal{M}$ to $\tilde{p}_\theta(z \mid s_t)$
        \State compute $F_\mathcal{M}, H_\mathcal{M}, C_\mathcal{M}$
        \State $U \gets w_F F_\mathcal{M} - w_H H_\mathcal{M} - w_C C_\mathcal{M}$
        \State $Q(\mathcal{M}) \gets (1-\alpha)Q(\mathcal{M}) + \alpha U$
    \EndFor
    \State $\mathcal{M}_t \gets$ random w.p.\ $\varepsilon$, else $\arg\max_\mathcal{M} Q(\mathcal{M})$
    \State \textbf{return} $\mathcal{M}_t$
\EndProcedure
\end{algorithmic}
\end{algorithm}

To couple update magnitude to execution quality, APGEM further modulates the learning rate by fidelity and entropy diagnostics, as summarized in Algorithm~\ref{alg:lr}.

\begin{algorithm}[h]
\caption{Policy Optimization with Adaptive Learning Rate}
\label{alg:lr}
\begin{algorithmic}[1]
\State \textbf{Input:} parameters $\theta$, base rate $\eta_0$, scale $\beta$, discount $\gamma$, weights $(\kappa,\lambda)$
\For{episode $=1$ to $N$}
    \State roll out trajectory $\tau$ under mitigated policy $\pi_\theta^{\mathcal{M}_t}$
    \State compute returns $G_k \gets \sum_{j=k}^{|\tau|-1}\gamma^{\,j-k} r_j$
    \State $\eta_t \gets \eta_0\,(1 + \kappa F_{\mathcal{M}} - \lambda H_{\mathcal{M}})$
    \State $\theta \gets \theta + \eta_t\,\nabla_\theta J(\theta;\mathcal{M}_t)$; wrap $\theta \gets \theta \bmod 2\pi$
\EndFor
\State \textbf{Output:} optimized parameters $\theta^\star$
\end{algorithmic}
\end{algorithm}

\begin{figure}[htbp!]
    \centering
    \includegraphics[width=0.5\linewidth]{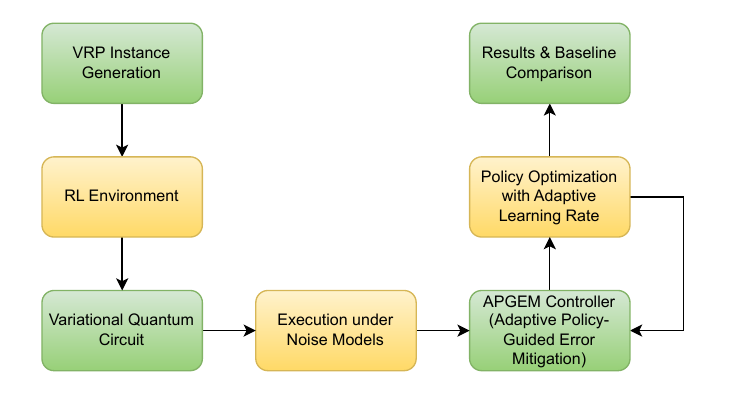}
    \caption{Proposed framework. VRP instance generation feeds an RL environment whose policy is a variational quantum circuit. Execution under noise is corrected online by the APGEM controller, with fidelity and entropy diagnostics guiding policy optimization; solutions are compared against classical baselines.}
    \label{fig:framework}
\end{figure}

\section{Performance Metrics}
We evaluate along three complementary layers. \textbf{Quantum reliability} is captured by state fidelity $F(\rho_\text{noisy},\rho_\text{ideal})$, the overlap between noisy and ideal states, and the normalized von Neumann entropy $H(\rho)=-\mathrm{Tr}(\rho\log_2\rho)$ divided by $n_q$, where lower entropy indicates a purer, more reliable execution. \textbf{Optimization effectiveness} is measured by cumulative reward $R=\sum_t r_t$, total routing cost $C$, and the approximation ratio $\alpha=C_\text{quantum}/C_\text{classical}$, where $\alpha$ near one indicates parity with the classical baseline. \textbf{Robustness} is quantified by the mitigation gain $\Delta M = M_\text{mitigated}-M_\text{unmitigated}$ for a metric $M$, the selection frequency of each mitigation technique, and the variance of reward, cost, and fidelity across seeds. Together these ensure that the evaluation reflects not only solution quality but also quantum reliability and adaptive resilience.

\section{Experimental Setup}\label{sec:setup}
We deliberately ground the evaluation in a real urban geography rather than a synthetic point set, so that the routing task reflects the spatial structure of an operational last-mile scenario. Instances use a single depot at Connaught Place (Delhi) with customer nodes drawn from real landmarks; inter-node costs are geodesic great-circle distances, preserving the true metric relationships between locations. Demands are sampled with zero depot demand, and vehicle capacity is set as a function of total demand to ensure feasibility.

\begin{figure}[htbp!]
    \centering
    \includegraphics[width=0.9\linewidth]{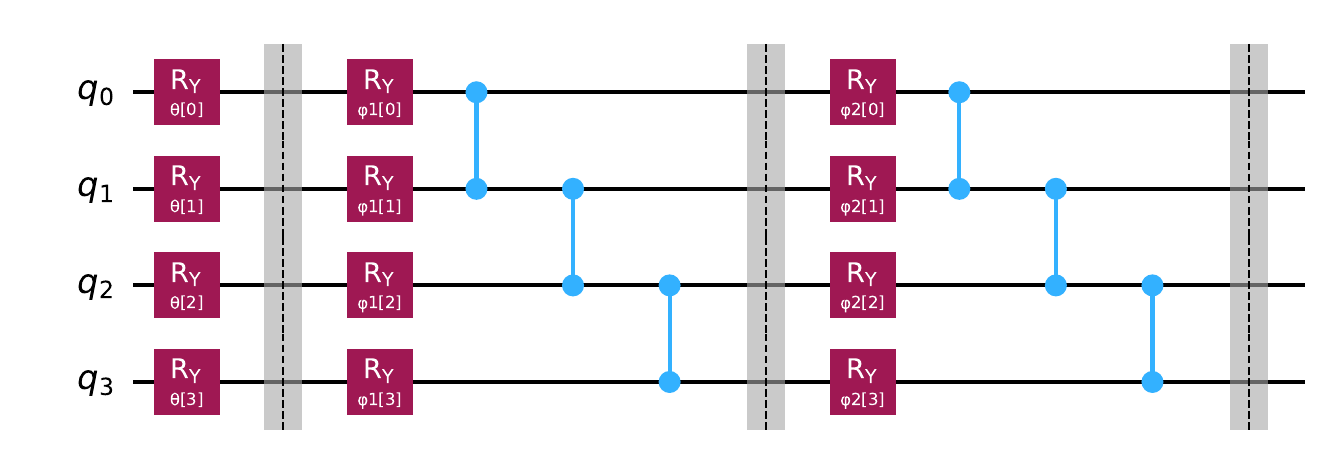}
    \caption{Quantum policy ansatz. Feature-encoding $R_y$ rotations embed the environment state, a controlled-$Z$ layer entangles qubits, and stacked variational $R_y$ layers provide trainable expressivity before measurement drives action selection.}
    \label{fig:ansatz}
\end{figure}

The policy ansatz, shown in Fig.~\ref{fig:ansatz}, applies $R_y(\theta_i)$ feature-encoding rotations, a controlled-$Z$ entanglement layer, and stacked trainable $R_y(\phi_i^{(k)})$ variational layers, balancing expressivity against circuit depth for NISQ feasibility. Circuits are simulated on the AerSimulator, for which the noiseless reference state is directly accessible; the state fidelity used in APGEM's utility is therefore computed exactly against this ideal reference at negligible cost. On real NISQ hardware the exact ideal state is not available, so the fidelity term would instead be estimated through hardware-computable surrogates (for example, near-Clifford or mirror-circuit fidelity proxies), or replaced by the entropy and estimator-variance diagnostics that the controller already tracks; the APGEM selection logic is unchanged in either case. Hardware-side fidelity estimation is left to future work. To emulate NISQ conditions, circuits execute under five noise families (depolarizing, amplitude damping, phase damping, two-qubit gate, and readout) at severity levels $\{0.01, 0.05, 0.08, 0.10\}$, yielding a grid of scenarios. Classical baselines are Nearest Neighbor, Savings (Clarke--Wright), Genetic Algorithm, Simulated Annealing, and Tabu Search. Training horizons of 100 and 500 episodes probe initial learning and stabilization.

\textbf{Implementation details.} The CVRP instance comprises a single depot, $8$ customer nodes, and a fleet of $3$ vehicles, with customer demands sampled randomly. The policy is encoded on $4$ qubits: the ansatz applies an $R_y$ feature-encoding layer, a controlled-$Z$ entanglement layer, and a RealAmplitudes variational block of two stacked trainable $R_y$ layers, giving $12$ trainable parameters ($4$ qubits $\times$ $3$ rotation layers). Circuit parameters are initialized uniformly at random in $[0,2\pi)$, and the discount factor is $\gamma=1$ (episodic). Reported metrics are averaged over multiple random seeds, with the same seeds shared across noise families and mitigation settings to enable paired comparison.

\section{Results}
 Table~\ref{tab:qrl_vs_classical} reports QRL against classical baselines. The agent outperforms constructive heuristics and approaches metaheuristics; against Nearest Neighbor the approximation ratio averages $0.743$ with a success rate above $92\%$, while ratios against metaheuristics lie between $0.83$ and $0.91$. Extending training to 500 episodes improves ratios uniformly, for example Genetic Algorithm from $0.909$ to $0.915$ and Tabu Search from $0.890$ to $0.896$, with reward curves converging $15$ to $20\%$ above the unmitigated baseline and approximation ratios stabilizing between $0.92$ and $0.94$ at lower variance.

\begin{table}[htbp!]
\centering
\caption{QRL vs.\ classical heuristics (100 vs.\ 500 episodes). Ratio is QRL/heuristic; success in \%.}
\label{tab:qrl_vs_classical}
\begin{tabular}{@{}lccccc@{}}
\toprule
\textbf{Method} & \textbf{Cost} & \multicolumn{2}{c}{\textbf{100 ep.}} & \multicolumn{2}{c}{\textbf{500 ep.}}\\
\cmidrule(lr){3-4}\cmidrule(lr){5-6}
 & & Ratio & Succ. & Ratio & Succ.\\
\midrule
Nearest Neighbor    & 108.29 & 0.743 & 92.2 & 0.748 & 92.1\\
Savings             & 90.57  & 0.888 & 57.6 & 0.894 & 57.9\\
Genetic Algorithm   & 88.52  & 0.909 & 54.7 & 0.915 & 54.1\\
Simulated Annealing & 96.78  & 0.831 & 70.8 & 0.837 & 70.7\\
Tabu Search         & 90.36  & 0.890 & 57.2 & 0.896 & 57.6\\
\bottomrule
\end{tabular}
\end{table}

\textbf{Mitigation techniques.} Table~\ref{tab:mitigation} summarizes per-technique approximation ratios. Under the short horizon PEC is strongest at $0.922$, though its aggressive cancellation amplifies variance; ZNE, CDR, and REM cluster near $0.903$--$0.908$. Under the long horizon CDR overtakes PEC ($0.918$ vs.\ $0.913$), consistent with richer calibration data favoring regression-based correction, and all techniques yield positive improvement over the unmitigated baseline.

\begin{table}[htbp!]
\centering
\scriptsize
\setlength{\tabcolsep}{3pt}
\caption{Mitigation technique performance (100 vs.\ 500 episodes).}
\label{tab:mitigation}
\begin{tabular}{@{}lcccc@{}}
\toprule
\textbf{Technique} & \textbf{Ratio (100)} & \textbf{Impr. (100)} &
\textbf{Ratio (500)} & \textbf{Impr. (500)}\\
\midrule
ZNE & 0.903 & +0.5\% & 0.910 & +0.8\%\\
PEC & 0.922 & $-$1.6\% & 0.913 & +0.5\%\\
CDR & 0.907 & +0.1\% & 0.918 & +0.9\%\\
REM & 0.908 & $-$0.1\% & 0.914 & +0.7\%\\
\bottomrule
\end{tabular}
\end{table}

\begin{figure}[htbp]
    \centering
    \includegraphics[width=0.7\linewidth]{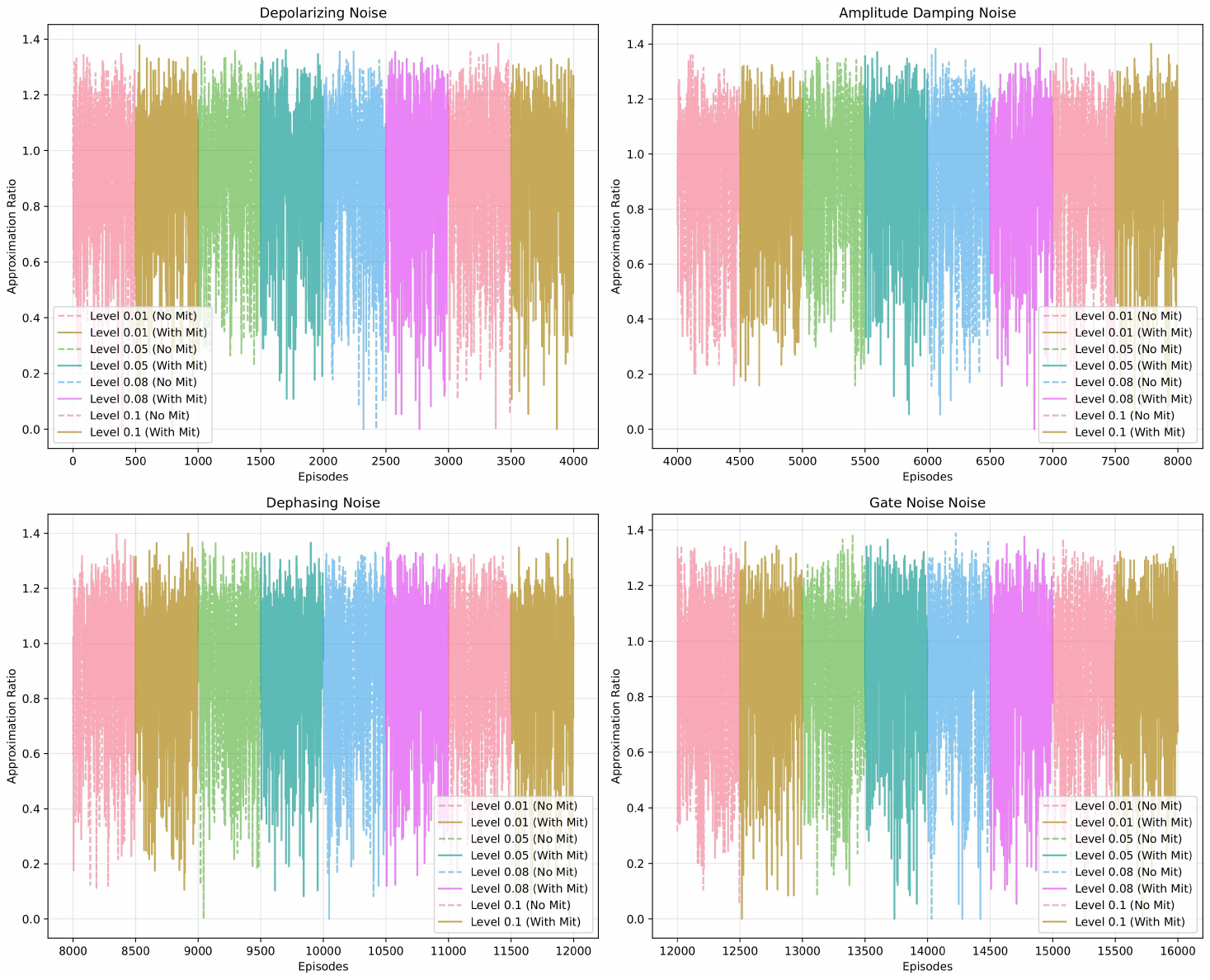}
    \caption{Noise sensitivity across five families over 500 episodes. Mitigation sustains approximation ratios above $0.92$ even at noise level $0.1$, well above the $0.84$--$0.87$ unmitigated range.}
    \label{fig:noise}
\end{figure}

\textbf{Noise resilience.} Table~\ref{tab:noise} and Fig.~\ref{fig:noise} report noise sensitivity. Across depolarizing, amplitude damping, dephasing, gate, and readout noise, unmitigated approximation ratios fall to $0.84$--$0.87$ at high noise ($0.08$--$0.10$), whereas mitigation restores them to $0.92$--$0.94$. Depolarizing, amplitude damping, and readout noise are mitigated most effectively; dephasing and gate noise remain partially challenging.

\begin{table}[htbp!]
\centering
\scriptsize
\setlength{\tabcolsep}{1.5pt}
\caption{Noise sensitivity and mitigation impact (100 vs.\ 500 episodes).}
\label{tab:noise}
\begin{tabular}{@{}lcccc@{}}
\toprule
\textbf{Noise} & \textbf{NM100} & \textbf{M100} & \textbf{NM500} & \textbf{M500}\\
\midrule
Depolarizing      & 0.86 & 0.93 & 0.87 & 0.94\\
Amplitude Damping & 0.85 & 0.92 & 0.86 & 0.93\\
Dephasing         & 0.84 & 0.91 & 0.85 & 0.92\\
Gate Noise        & 0.84 & 0.90 & 0.85 & 0.92\\
Readout           & 0.85 & 0.92 & 0.86 & 0.93\\
\bottomrule
\end{tabular}
\end{table}

\begin{figure}[htbp!]
    \centering
    \includegraphics[width=0.7\linewidth]{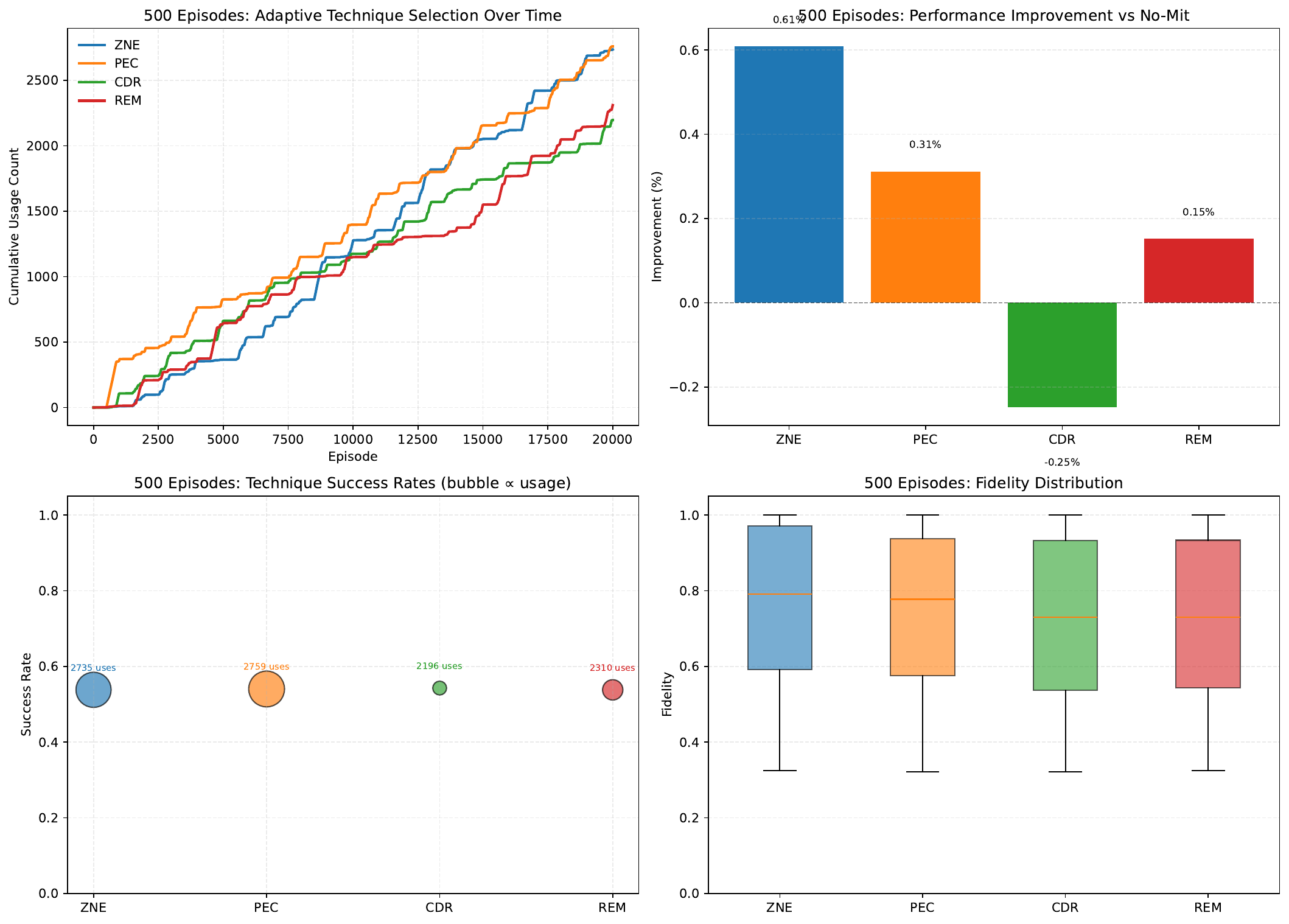}
    \caption{APGEM selection dynamics over 500 episodes. The controller distributes usage across ZNE, PEC, CDR, and REM as success rates converge, reflecting mature policy and mitigation co-adaptation.}
    \label{fig:adaptive}
\end{figure}

\textbf{Adaptive dynamics.} Table~\ref{tab:adaptive} and Fig.~\ref{fig:adaptive} report controller behavior, the central finding of this work. Under the short horizon the controller strongly favors CDR ($8{,}332$ selections), a conservative preference for calibration-based correction when data are scarce. Under the long horizon selections balance across ZNE ($24{,}067$), PEC ($24{,}485$), CDR ($19{,}305$), and REM ($20{,}692$), with success rates converging. This shift from a single dominant technique to diversified deployment indicates genuine regime-dependent selection and co-adaptation between policy learning and mitigation.

\begin{table}[htbp!]
\centering
\scriptsize
\caption{Adaptive mitigation dynamics (usage counts).}
\label{tab:adaptive}
\begin{tabular}{@{}lcccc@{}}
\toprule
\textbf{Tech.} & \textbf{U100} & \textbf{S100} & \textbf{U500} & \textbf{S500}\\
\midrule
ZNE & 3{,}498 & High & 24{,}067 & High\\
PEC & 3{,}884 & High & 24{,}485 & High\\
CDR & 8{,}332 & Mod. & 19{,}305 & High\\
REM & 1{,}962 & Mod. & 20{,}692 & High\\
\bottomrule
\end{tabular}
\end{table}

\section{Discussion}
Two patterns stand out. First, mitigation is not optional under realistic noise: at the highest severities it is the difference between sub-parity performance and approximation ratios near $0.94$. Second, the value of any single technique is regime-dependent. PEC leads under short horizons where its unbiased correction matters most, whereas CDR leads under long horizons as accumulated calibration data sharpen its regression. A static choice would lock in one of these regimes and forfeit the other; APGEM instead tracks the transition, which is visible as the shift from CDR-dominated selection to balanced deployment. Grounding the study in a real Delhi last-mile scenario shows the controller operating under the spatial structure of an operational routing task rather than a synthetic distribution, strengthening the practical reading of these results. The evaluation still centers on a single instance, which we treat as a focused case study; cross-instance benchmarking on standard suites is the natural next step and is already framed as future work.

\section{Conclusion}
We presented APGEM, an adaptive, learning-aware error mitigation controller for QRL applied to the CVRP. By selecting among ZNE, PEC, CDR, and REM online through a fidelity, entropy, and cost aware utility, APGEM restores approximation ratios under high noise and exhibits regime-dependent selection that matures from a CDR-dominated short horizon to balanced long-horizon deployment. Grounding the study in a real Delhi last-mile scenario demonstrates the controller under the spatial structure of an operational routing task, and these preliminary results support adaptive mitigation as a practical path toward noise-resilient QRL on NISQ hardware. Building on this case study, the framework extends naturally to standard CVRPLIB suites for cross-instance benchmarking, richer VRP variants, qubit-efficient encodings for larger instances, and deployment on real quantum hardware.


\end{document}